\documentclass[nonacm,sigconf]{acmart}

\renewcommand\footnotetextcopyrightpermission[1]{}

\usepackage{xcolor}
\usepackage{pifont}
\usepackage{hyperref}
\usepackage{cleveref}
\usepackage{multirow}
\usepackage{algorithm}
\usepackage{subfigure}
\usepackage{algpseudocode}
\usepackage{svg}
\usepackage{ulem}
\usepackage[flushleft]{threeparttable}
\usepackage{dcolumn}
\usepackage{arydshln}
\usepackage{multicol}

\begin{document}
\setlength{\aboverulesep}{-1pt}
\setlength{\belowrulesep}{0pt}
\setlength{\extrarowheight}{.85ex}
%

\title[]{CopyScope: Model-level Copyright Infringement Quantification in the Diffusion Workflow}

\author{Junlei Zhou}
\affiliation{%
  \institution{Southern University of Science and Technology}
  \country{China}
}
\email{12332458@mail.sustech.edu.cn}

\author{Jiashi Gao}
\affiliation{%
 \institution{Southern University of Science and Technology}
 \country{China}
}
 \email{12131101@mail.sustech.edu.cn}

\author{Ziwei Wang}
\affiliation{%
  \institution{Southern University of Science and Technology}
  \country{China}
}
 \email{12250053@mail.sustech.edu.cn}

\author{Xuetao Wei}
\authornote{Corresponding author.}
\affiliation{%
  \institution{Southern University of Science and Technology}
  \country{China}
}
 \email{weixt@sustech.edu.cn}


\begin{abstract}

Web-based AI image generation has become an innovative art form that can generate novel artworks with the rapid development of the diffusion model. However, this new technique brings potential copyright infringement risks as it may incorporate the existing artworks without the owners' consent. Copyright infringement quantification is the primary and challenging step towards AI-generated image copyright traceability. Previous work only focused on data attribution from the training data perspective, which is unsuitable for tracing and quantifying copyright infringement in practice because of the following reasons: (1) the training datasets are not always available in public; (2) the model provider is the responsible party, not the image. Motivated by this, in this paper, we propose \texttt{CopyScope}, a new framework to quantify the infringement of AI-generated images from the model level. We first rigorously identify pivotal components within the AI image generation pipeline. Then, we propose to take advantage of Fréchet Inception Distance (\textit{FID}) to effectively capture the image similarity that fits human perception naturally. We further propose the \textit{FID}-based Shapley algorithm to evaluate the infringement contribution among models. Extensive experiments demonstrate that our work not only reveals the intricacies of infringement quantification but also effectively depicts the infringing models quantitatively, thus promoting accountability in AI image-generation tasks.
  
\end{abstract}



\keywords{Accountability, Generative AI, Copyright traceability, Infringement}

\maketitle

\section{Introduction} 
\begin{figure}[t]
  \graphicspath{{./figures/}}
  \centering
  \includegraphics[width=1.0\linewidth, height=0.50\linewidth]{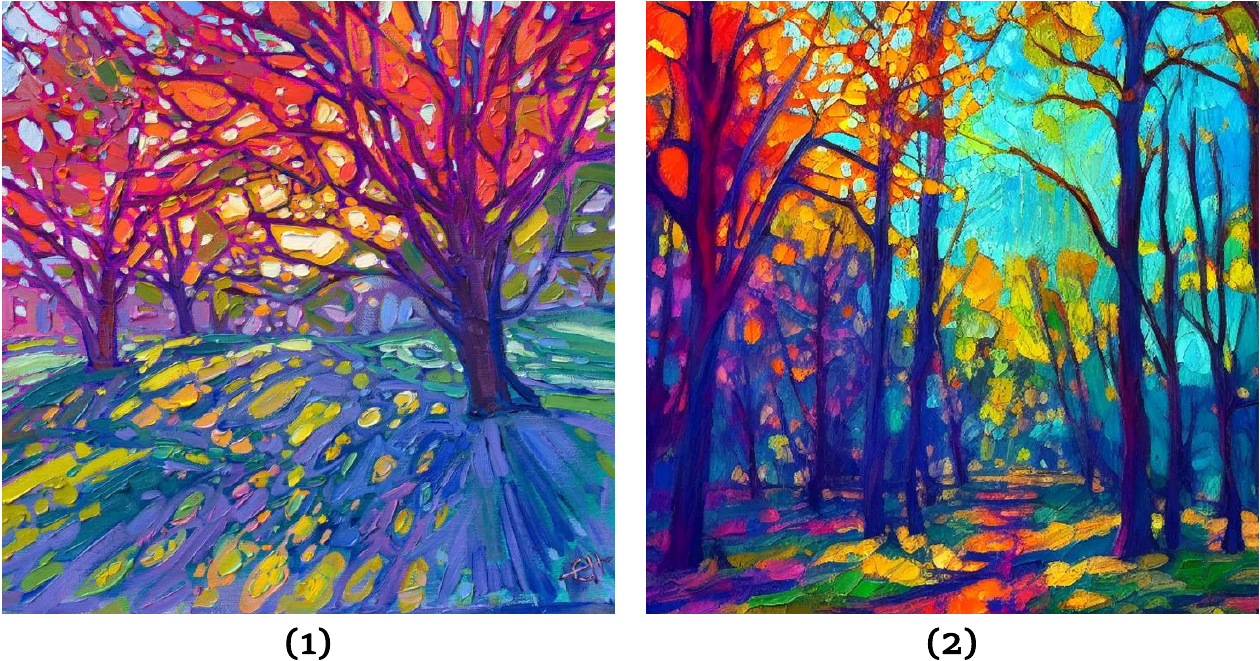}
  \caption{The image (1) was created by painter Erin Hanson in 2021, and the image (2) was created by Stable Diffusion using the "style of Erin Hanson" as a prompt. The styles of these two images are so similar that it is impossible to tell them apart. }
  \Description{Oil painting on crystal, light and shadow, back lit tree, strong silhouette, stained glass, modern impressionism, style of Erin Hanson}
  \label{figure:1}
\end{figure}

AI image generation programs, namely Artificial Intelligence Generated Content (AIGC) tools such as Stable Diffusion \cite{stablediffusion}, DALL·E2 \cite{DallE2}, Midjourney \cite{Midjourney}, are setting off a new revolution of artwork creation. These programs allow users to effortlessly generate target images by taking some descriptions as input into the model. However, these AI-generated artworks inherit the characteristics of the images that are used to train models \cite{wang2023evaluating, 8695364, sha2022fake}, which might be pretty similar to the original ones, as shown in Figure \ref{figure:1}. Such similarity has aroused concerns about copyright infringement disputes. For example, three artists (Sarah Andersen, Kelly McKernan, and Karla Ortiz) have recently accused Stable Diffusion of unlawfully scraping copyrighted images from the Internet to mimic their art styles\cite{GettyImages}. To this end, research on responsible AI image generation is in urgent need to address such copyright issues.

\begin{figure}[t]
  \centering
  \includegraphics[width=1.0\linewidth, height=0.92\linewidth]{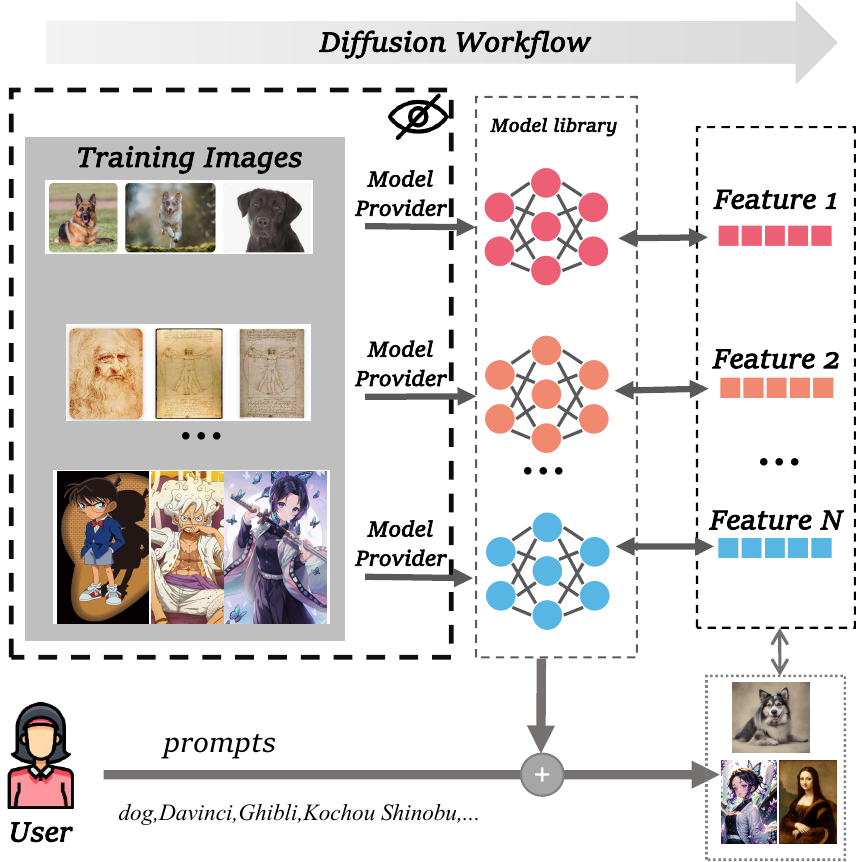}
  \caption{The interaction between user and model. It shows that the training images are invisible other than the model provider, and each model is uniquely associated with its provider, which is possible to conduct a model-side infringement tracing.}
  \Description{}
  \label{figure:3}
\end{figure}

Previous work on data attribution \cite{Mahmood_2022_CVPR, jia2021scalability, kwon2021beta, sim2022data} focused on how the images in the training data contributed to the model's outcome, which is not suitable for the context of copyright traceability. This is because of the following reasons: (1) the training data is not known in advance in real-world practices; as shown in Figure \ref{figure:3}, the models are usually publicly available, while the training datasets are not\cite{wang2023evaluating, dai2023training}. (2) the responsible party is the model provider, namely, the personnel or the organization who abused the online image collections without the owners' consent, not the image itself. As long as the model that generates the infringing image is identified, the corresponding infringer (the model provider) can be found, and the degree of infringement can be quantified. Thus, we need to develop a new approach to quantify copyright infringement from the model level, which is the focus of our paper.

To this end, we propose a new framework \texttt{CopyScope} at the model level towards AIGC copyright traceability. Our framework \texttt{CopyScope} includes three closely intertwined stages (\texttt{Identify-} \texttt{Quantify-Evaluate}). In the \texttt{Identify} stage, we conduct an extensive and in-depth analysis of 16,000 generated images under six themes and rigorously identify four components (\texttt{Base Model}\cite{9878449}, \texttt{Lora}\cite{hu2021lora}, \texttt{ControlNet}\cite{zhang2023adding}, and \texttt{Key Prompt}\cite{pmlr-v139-radford21a}) that are involved in the infringement in diffusion workflow. In the \texttt{Quantify} stage, we extensively compare and analyze five metrics, which include \textit{Cosine}\cite{cosine}, \textit{DHash} (Difference Hash similarity)\cite{dHash}, \textit{Hist} (Histogram similarity)\cite{hist}, \textit{SSIM} (Structural similarity)\cite{wang2004image}, and \textit{FID} (Fréchet Inception Distance)\cite{heusel2017gans} methods,  from multiple dimensions, such as style, structure, etc., to measure the similarity between the generated and original images. We find that the \textit{FID} metric is the most effective quantification method because \textit{FID} can capture the similarity that fits human perception naturally, which could reflect accurate quantification of each model's contribution. In the \texttt{Evaluate} stage, we model our scenario into a cooperative game model and propose the \textit{FID}-based Shapley method to evaluate the contribution of each infringement model. In the end, we conduct extensive experiments to demonstrate that our proposed \textit{FID}-based Shapley algorithm could effectively quantify the infringement in the diffusion workflow, which offers a promising approach to help us unveil the intricacies of infringement in the emerging domain of AI image generation tasks. 

We summarize our contributions as follows:
\begin{itemize}
    \item To our knowledge, we are the first to propose a new copyright infringement quantification framework \texttt{CopyScope} at the model level, which facilitates stakeholders to investigate the emerging intricate infringement case. 
    \vspace{1mm}
    \item We propose the \textit{FID}-based Shapley algorithm to effectively quantify the infringement in the diffusion workflow, which takes advantage of Fréchet Inception Distance (\textit{FID}) to effectively capture the image similarity that fits human perception naturally, and the Shaplely value scheme to quantify the infringement contribution among models.
   \vspace{1mm}
    \item We conduct extensive experiments to demonstrate the effectiveness of our proposed framework \texttt{CopyScope}, which depicts the infringing models quantitatively, thus promoting the legally compliant use of AI-generated content.
    
\end{itemize}

\begin{figure*}
  \includegraphics[width=\textwidth]{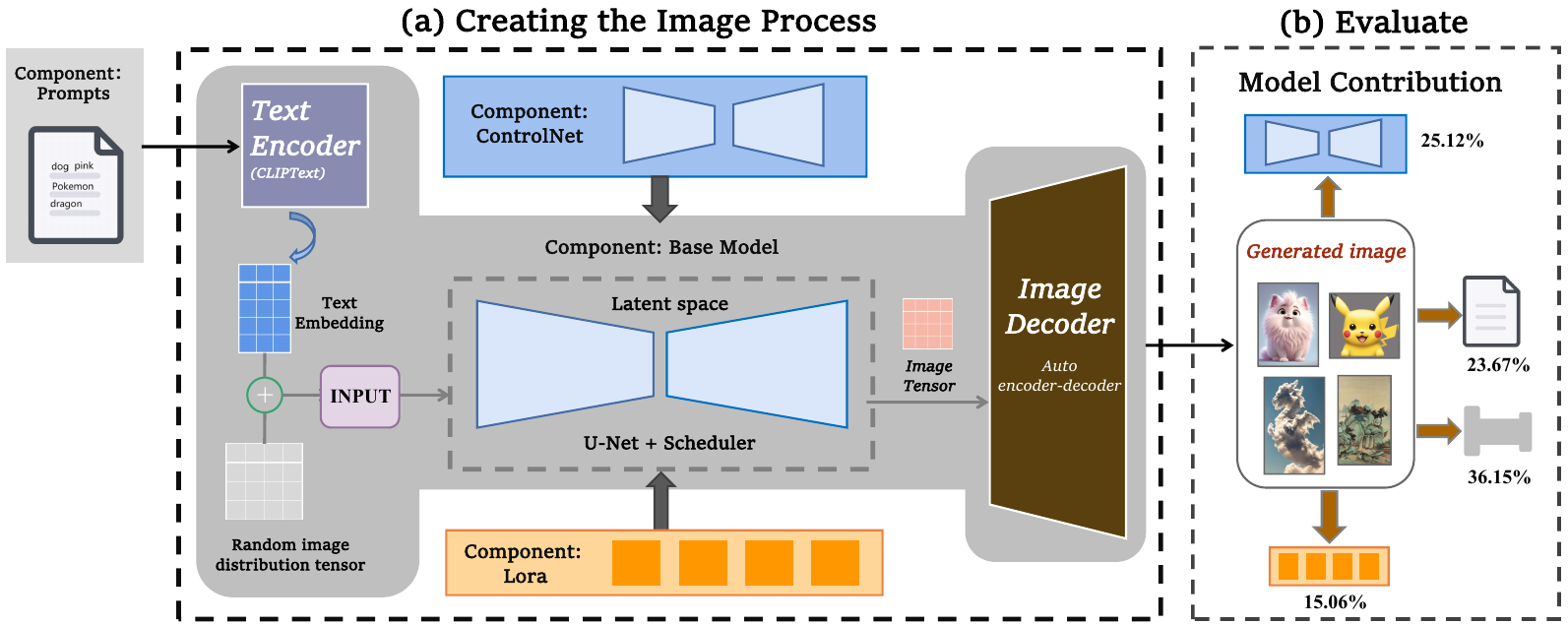}
  \caption{
  Illustration of the diffusion workflow, which consists of two parts: (a) the image creation process, in which multiple components jointly contribute to the generation of images; (b) the Evaluate stage, in which the contribution of multiple models on the generated image's copyright is evaluated.}
  \label{figure:2}
\end{figure*}

\section{Preliminary and Background}
\subsection{AI Image Generation}
In Figure~\ref{figure:2}, we show the basic diffusion workflow to generate images with text as input:
\ding{182} First, a user sets up a fundamental component to create images, which includes the text encoder model, U-Net $\&$ scheduler, and image decoder.
\ding{183} The user selects specific prompts and adjusts related parameters such as seed, sampling steps, image size, etc. 
However, due to the universality of the fundamental component, the above process alone can merely generate coarse images. Therefore, additional components such as \textit{low-rank adaptation} of large language models (Lora)\cite{hu2021lora} and \textit{conditional control} of text-to-image diffusion models (ControlNet)\cite{zhang2023adding} are required to create fine images with specific themes. These components jointly contribute to the output image’s characteristics, including content, style, and background, which refer to the originality of the image and pose a potential risk of copyright infringement. The functionality of each component is described as follows:

\begin{enumerate}
    \item \textbf{\texttt{Base model}}: The most famous model for AI image generation is the SD series model (e.g., SD1.5\cite{9878449}, SD2.0, and SD XL\cite{SD-XL}.) released by StabilityAI\cite{stabilityai} and Runway \cite{runwayml}. However, as these models are trained using hundreds of millions of images with mixed styles, they are difficult to generate images with a specific theme. Therefore, the image generation usually relies on secondary-trained models based on these basic models (e.g., DreamShaper\cite{dreamshaper}, GhostMix\cite{ghostmix}, Anything\cite{anything}, SDMv10\cite{SDMv10}). These secondary-trained models have unique themes and styles and are more likely to infringe the existing works. We thereby conduct copyright traceability based on the secondary-trained model and refer to it as the \texttt{Base Model}. 
    
    
    \item \textbf{U-Net $\&$ Scheduler}: U-Net iteratively denoises the Gaussian noise matrix in a diffusion loop, and the noise of each prediction is guided by text and timesteps. The predicted noise is removed from the random Gaussian noise matrix, and finally, the random Gaussian noise matrix is converted to the latent features of the image. The scheduler is responsible for the forward and backward propagation of the entire Diffusion model. It processes the model's output during training and inference according to the set mathematical rules and the number of timesteps.
    
    
    \item \textbf{\texttt{Lora}}: A fine-tuning model generates a specific type of image with a particular style of artist (e.g., Ghibli Style\cite{Ghibli}, Davinci Style\cite{Leonardo}, One Piece Style\cite{onepiece}). Such models are trained at lower cost and are more likely to generate images with particular topics, making them more susceptible to producing infringing images. 
    
    \item \textbf{\texttt{ControlNet}}: A neural network model is deployed to regulate the stable diffusion model, which enables control over the primary structure of the generated image. We can transfer the composition or human pose from the reference image to the target image by utilizing \texttt{ControlNet}. Furthermore, \texttt{ControlNet} is almost inseparable in some specific generation tasks, such as specific image structure, design layout, architectural form, line drawing coloring, and other scenes\cite{Depth}.  
    
    \item \textbf{\texttt{Prompt}}: Users can leverage their imagination and employ appropriate prompt words to describe their images to AI to generate desirable results. Based on the content, description instructions can be categorized into type prompts, content prompts, composition prompts, and painter prompts. Among these, painter prompts are primarily intended for generating the image's subject, which can be an artist's painting style or the protagonist of a specific work. However, it is essential to note that such prompts often directly infringe upon the original author's rights, which are called \texttt{Key Prompt} in our research.
\end{enumerate}




\begin{table*}
    \centering
    \begin{tabular}{c | c | l}
    \toprule[1.5pt]
         \textbf{Components} & \textbf{Description} & \textbf{Models} \\
    \hline
	\multirow{2}{*}{\texttt{Base Model}} & \multirow{2}{5.3cm}{The basic model of stable diffusion determines the style of generated image.} & \textit{SDv1-5}\cite{SDv1-5}: Universal model without specific topic. \\           \cline{3-3}
                      &  & \textit{SDMv10}\cite{SDMv10}: Based on \textit{SDv1-5} and added training of classical works of art. \\ 
        \hline
        \multirow{2}{*}{\texttt{ControlNet}} & \multirow{2}{5.3cm}{A category of models that control image\\ structure by adding additional conditions.} & \textit{Depth}\cite{Depth}: Capture the original image's structural depth to control the\\ 
                                             &  & generated image's structure. \\
        \hline
        \multirow{2}{*}{\texttt{Lora}} & \multirow{2}{5.3cm}{Fine-tune the generated image.} & \textit{Leonardo}\cite{Leonardo}: Use Davinci's portfolio training to adjust images to more \\ 
                                       & & closely resemble Davinci's creative style.\\
        \hline
        \multirow{2}{*}{\texttt{Key Prompt}} & \multirow{2}{5.3cm}{Instruct an AI on what to paint.} &  \textit{Davinci}: Tips for generating Leonardo da Vinci style images. \\  \cline{3-3}
                                             & & \textit{MonaLisa}: Make the generated image closer to MonaLisa. \\      
    \bottomrule[1.5pt]
    \end{tabular}
    \caption{Four infringement components have been identified, each representing a type of model set. A brief description of their function is in the Description column. The Models column shows the specific models we used in the Mona Lisa experiment (please see \cref{experiment}.)}
    \label{tab:1}
\end{table*}
\subsection{Similarity Metrics}
\label{sec:similaritymetrics}
To study the issue of copyright tracing more accurately and effectively, we need to quantify the models involved in infringement. Regarding AI-generated images, we need to examine their relationship with the original image from multiple dimensions, such as style, structure, and other characteristics. Therefore, we quantify the models associated with infringement based on these dimensions and compare the following quantitative indicators: 

\textbf{Cosine similarity (\textit{Cosine})}. Cosine similarity calculates the cosine of the angle between the vectors, equivalent to the vectors' dot product divided by the product of their lengths. In the image similarity calculation,
we can convert the image into a feature vector and then use cosine
similarity to compare the similarity of these feature vectors. The image cosine similarity ranges from $[0,1]$, where $1$ indicates the same vectors and $0$ indicates orthogonality or decorrelation.

\textbf{Difference Hash similarity (\textit{DHash})}. The essence of the hash algorithm for image similarity recognition is to hash the image to generate a set of binary numbers and then find similar images by comparing the hash value distance of different images. \textit{DHash}\cite{dHash} is a differential hashing algorithm that compares the size of the left and right pixels when hashing an image to obtain the hash sequence. A larger hash value indicates a more tremendous difference between the two images.

\textbf{Histogram similarity (\textit{Hist})}. It measures the similarity of two pictures in color distribution. The histogram algorithm counts the number of pixels of different colors in the image, presents it as a histogram, and then compares the image similarity. The value range is $[0,1]$. The closer to $1$, the more similar the two are. However, the histogram method only considers the color distribution but ignores texture and structure information.

\textbf{Structural similarity (\textit{SSIM})}. \textit{SSIM}\cite{wang2004image} simultaneously compares the similarity of images from three aspects: brightness, contrast, and structure. The \textit{SSIM} algorithm can enhance the structural similarity of images in a group and better detect subtle differences.
    \begin{equation}
    \label{Eq:9}
    \textbf{SSIM}(r, g) = \frac{{(2\mu_r\mu_g + C_1)(2\sigma_{rg} + C_2)}}{{(\mu_r^2 + \mu_g^2 + C_1)(\sigma_r^2 + \sigma_g^2 + C_2)}}.
    \end{equation}
Among them, \(\mu \) represents the average brightness of the image,\(\sigma \) represents the standard deviation of image brightness, \(\sigma \) represents the variance of image \(r\), \(g\), and \(C\) is a constant. The value range of \textit{SSIM} is \([0,1]\). The larger the value, the more similar the two images are.

\textbf{FID method (\textit{FID})}. Fréchet Inception Distance\cite{heusel2017gans}, originally used as an evaluation index of the generative model to calculate the distance between the real image and the generated image feature vector. \textit{FID} directly considers the distance between the generated data and the accurate data at the feature level, and the smaller the data, the more similar. From a theoretical perspective, \textit{FID} measures the distance between two multivariate normal distributions, and its calculation formula is as follows,
    \begin{equation}
    \label{Eq:10}
    \textbf{FID} = {\parallel \mu_r - \mu_g \parallel}_2^2 + Tr(\Sigma_r + \Sigma_g - 2(\Sigma_r \Sigma_g)^{1/2}).
    \end{equation}
Among them, \(\mu_r\) and \(\mu_g\) represent the mean value of the feature vectors extracted from the real image and the generated image, respectively; \(\Sigma_r\) and \(\Sigma_g\) represent the covariance matrix of the original image and the generated image.

\begin{table*}[ht]
\vspace{-1.0em}
    \centering
    \begin{tabular}{c|cccccc}
        \toprule[1.5pt]
         &  \textbf{Celebrity} &  \textbf{Film$\&$TV} & \textbf{Artwork} & \textbf{Popular models} &  \textbf{Design} &  \textbf{Game} \\
         \hline
         \texttt{Base Model} & 3,164(100$\%$) & 3,210(100$\%$) & 3,031(100$\%$) & 3,383(100$\%$) & 1,962(100$\%$) & 1,711(100$\%$)\\
         \texttt{ControlNet} & 1,526(48$\%$)& 1,037(32$\%$) & 1,196(39$\%$)  & 1,338(39$\%$) & 352(17$\%$) & 576(33$\%$)\\
         \texttt{Lora} & 2,959(93$\%$) & 2,893(90$\%$) & 2,167(71$\%$) & 2,764(81$\%$) & 1,158(59$\%$) & 1,711(100$\%$)\\
         \texttt{Key Prompt} & 3,079(97$\%$) & 2,574(80$\%$) & 2,386(93$\%$) & 1,996(59$\%$) & 1,286(65$\%$) & 1,077(62$\%$)\\
         \bottomrule[1.5pt]
    \end{tabular}
    \caption{The dataset for identifying infringement components in the \texttt{CopyScope} framework. The generated images in the dataset come from AI-generated images platform \texttt{Civitai}\cite{civitai}. The generated images in the dataset are divided into six major themes, and the usage ratio of the four types of infringement components in each category is marked.}
    \label{tab:2}
\end{table*}

\section{Methodology}
In this section, we present an \texttt{Identify-Quantify-Evaluate} framework called \texttt{CopyScope} to address the issue of AI-generated image copyright traceability at the model level. In the \texttt{Identify} stage, we first rigorously select four pivotal components for describing infringing models by analyzing images generated from \texttt{Civitai}\cite{civitai}. In the \texttt{Quantify} stage, we adopt \textit{FID} to measure the similarity between the original images and the images generated by our designed model under different alliances of models. In the \texttt{Evaluate} stage, we trace back the possible infringing model by computing the contribution of models using the \textit{FID}-Shapley value.

\subsection{Identify Influential components}
We initiate our study by determining components that have the most significant impacts on the generated images. Components make up the AI image generation workflow, which is used to characterize generation models in our proposed copyright tracking approach.  

This stage is based on a survey from the world’s largest AI image generation exchange and sharing platform \texttt{Civitai}, where we collect more than 16,000 generated image data from over 5,000 models to find commonalities in generated images. The generated images are divided into 6 themes: celebrity, film\&TV, artwork, popular models, design, and game. We explore the distribution of models that generate images involving copyright infringement by calculating the usage rate of components: \texttt{Base Model}, \texttt{Lora}, \texttt{ControlNet}, and \texttt{Key Prompt}. Table~\ref{tab:2} shows the frequency of these components, indicating that they have a high usage rate in AI image-generation tasks.

We identify four components that are used in AI image generation at a high frequency: \ding{182} The \texttt{Base Model} is essential for each generated image. \ding{183} The second is the \texttt{Lora}. Although the \texttt{Lora} is not a necessary option for generating images, we can find from Table~\ref{tab:2} that it has a high application rate in each category, indicating that the use of \texttt{Lora} for adjustment in AI image generation has become a norm. \ding{184}  Prompts with particular specificity are called \texttt{Key Prompts}. \texttt{Key Prompt} can make the generated image close to the characteristics of these keywords to a large extent, thus infringing on the original author's rights. \ding{185} The overall usage rate of \texttt{ControlNet} is relatively lower than the other components. This is because the \texttt{ControlNet} is challenging to use as it needs higher environment configuration requirements than \texttt{Lora} and \texttt{Key Prompt}\cite{extensions}. However, \texttt{ControlNet} is an essential components in generating particular themes as it controls the structure of the image. From the perspective of copyright tracing, the \texttt{ControlNet} is a critical suspected infringement component that our \texttt{CopyScope} framework considers.

\subsection{Quantify Model Performance}
As shown in Table~\ref{tab:1}, we choose specific models for each component. We select \textit{SDv1-5} and \textit{SDMv10} for \texttt{Base Model}, \textit{Depth} for \texttt{ControlNet}, \textit{Leonardo} for \texttt{Lora}, \textit{Davinci} and \textit{MonaLisa} for \texttt{Key Prompt}. We use these models to simulate 30 different alliances, where the two specific models of \texttt{Base Model} is essential for any alliances of models and the other four specific models of  \texttt{ControlNet}, \texttt{Lora}, \texttt{Key Prompt} can be used to form an alliance at the same time. In this way, the alliance of models is calculated as $2\times(C_{4}^{1} + C_{4}^{2} + C_{4}^{3} + C_{4}^{4})=30$. We then use each alliance to generate $100$ batches of images to generate the dataset of $3,000$ images for quantification and evaluation. 

We propose to explore which model can affect the generated image and thereby cause infringement by studying the generated and original images. Therefore, accurately measuring the similarity between the generated and original images is crucial to quantification. We quantify the performance of each alliance by measuring the similarity of the images generated by the alliance and the original image. To this end, we present the results of multiple indicators such as \textit{Cosine}, \textit{Hist}, \textit{DHash}, \textit{SSIM}, and \textit{FID} as described in \cref{sec:similaritymetrics}.

According to the results in the following experiment, we comparatively select \textit{FID} as the quantification method because the results of images' similarity are more approximate to the human intuition. Furthermore, in \cref{quantifyexperiment}, we compare these indicators and found that \textit{FID} has a more accurate similarity description ability than other quantitative methods. Additionally, the discrimination between the generated results of different alliances is more remarkable, which is more helpful for our subsequent evaluation.

\subsection{Evaluate Contributions of Models}
\label{sec: evaluation_methods}
In the sections above, we have identified the copyright-related components in the diffusion workflow and proposed quantitative metrics to measure image similarity. To achieve copyright traceability and determine the level of infringement for different models, we introduce the \texttt{Evaluate} stage as the final step of the \texttt{CopyScope} framework, as shown in Figure \ref{figure:2}. The \texttt{Evaluate} stage aims to assess the contribution of each model in the output image. We formally define the \texttt{Evaluate} stage as follows:
\begin{definition} (\texttt{Evaluate} stage) 
Given a set of models $\mathcal{M} = \{z_1, z_2, \dots, z_n\}$, where each $z_i$ is a specific model of the component in the diffusion workflow, the \texttt{Evaluate} stage aims to find the value of $v(z)$ of each model $z$, which represents how much it contributes to the copyright of the generated image.
\end{definition} 

Let $\mathcal{L}=\left \{z_1,...,z_N  \right \} $ be an alliance formed by $N$ models and $U(\cdot)$ be the value function that can be applied on any subset of the alliance $\mathcal{L}$. To evaluate the contribution of $i$-th model $z_i$ to the overall value of the alliance $\mathcal{L}$, two widely used solutions from cooperative game theory are Leave-one-out (LOO) \cite{ying2004fast} and Shapley value (SV) \cite{ijcai2022p778}.

\noindent\textbf{LOO-based Method.} The idea of  LOO is to measure the marginal contribution of each model $z$ to the alliance by removing it from the alliance and observing the difference. For the $i$-th model $z_i$, its contribution $v_{LOO}(z_i)$, also referred to as the LOO value, can be obtained as follows,
\begin{equation} 
   v_{LOO}(z_i) \propto U(\mathcal{L})-U(\mathcal{L}\backslash {z_i}). 
  \label{LOO:formula} 
\end{equation}

\noindent \textbf{SV-based Method.}
The SV-based method was originally used to provide a fair way of dividing the benefits for players in a coalition based on their individual and joint contributions. To calculate the contribution of  a model in the alliance,  the SV-based method considers all possible sub-alliances that include that model and then takes the weighted average of the differences between the value of each sub-alliance with and without that model. For the $i$-th model $z_i$, its contribution $v_{SV}(z_i)$, also referred to as the Shapley value, can be obtained as follows,
\begin{equation}
    v_{SV}(z_i) \propto \frac{1}{N}\sum_{\mathcal{L} \subseteq \mathcal{M} \setminus {z_i}} [U(\mathcal{L} \cup {z_i})-U(\mathcal{L})] \frac{| \mathcal{L} |!(N-1-|\mathcal{L}|)!}{(N-1)!}.
    \label{Shapley:formula}
\end{equation} 

\begin{algorithm}[t]
\small
    \caption{FID-Shapley Algorithm}
    \begin{algorithmic}[1]
    \State{\textbf{Input:} All models: $\mathcal{M} =\left \{ z_1,...,z_N \right \} $; the alliances set $\mathcal{S} = \{\mathcal{L}_1, \mathcal{L}_2,\dots, \mathcal{L}_n\}$, where $\mathcal{S}$ contains all possible alliance $\mathcal L$ from  $\mathcal{M}$; the \textit{FID}-based value function $U(\cdot)$}.
    \State {\textbf{Initialize}: An \textit{FID}-Shapley value set $\mathcal R = \mathcal \varnothing$,  \textit{FID}-Shapley value $v=0$}.
    \For {each model $z$ in $\mathcal{M}$} 
        \State{Initialize margin contribution $ c(z) = 0$;}
        \For{each alliance $\mathcal{L}$ in $\mathcal{S}$}{
            \If{$z \notin \mathcal{L}$} 
                \State{Continue;}  
            \Else 
                \State{Update the margin contribution of model $z$:}
                
                \State{$c(z) \gets c(z)+ [U(\mathcal{L})-U(\mathcal{L}-z)] \times \frac{| \mathcal{L} |!(N-1-|\mathcal{L}|)!}{(N-1)!} $;}

            \EndIf
            }
        \EndFor 
        \State{Calculate the \textit{FID}-Shapley value of $z$: $v(z) = \frac{c(z)}{N}$;} 
        \State{Append to \textit{FID}-Shapley value set $\mathcal{R} \leftarrow  v(z)$;}
    \EndFor
    \State {\textbf{Return:} $\mathcal{R}$.}
    \end{algorithmic}
    \label{Algorithm:1}
\end{algorithm}
We represent the models in the diffusion workflow as an alliance in a cooperative game\cite{pmlr-v97-ghorbani19c, 10.1145/3450439.3451861, 10.1007/978-3-030-86137-7_15}, as each model jointly forms an image generation alliance, as shown in Table \ref{tab:3}, and the value of the model can be calculated by a value function, which is the similarity metrics in our work.  Each model is independent and does not affect each other, and the diffusion workflow can be built by any sub-alliance from the models. Meanwhile, the models' alliance satisfies the following properties\cite{shapley1951notes}, which enables us to design a contribution evaluation method based on the Shapley value.

\begin{table*}[t]
\centering
  \begin{tabular}{llccccccc}
    \toprule[1.5pt]
    \textbf{Figure No.} & \textbf{Alliances }&  \textbf{\textit{Cosine}} $\uparrow$ & \textbf{\textit{Hist}}$\uparrow$ & \textbf{\textit{DHash}}$\uparrow$ & \textbf{\textit{SSIM}}$\uparrow$ & \textbf{\textit{RGB-SSIM}} $\uparrow$ & \textbf{\textit{FID}}$\downarrow$\\
    \midrule
    Figure~\ref{figure:7}(1) & \textit{SDv1-5} & 0.8802 & 0.1511 & 0.5468 & 0.2934 & 0.8424 & 310.18 \\
    Figure~\ref{figure:7}(2) & \textit{SDv1-5}+\textit{Depth} & 0.8927 & 0.3518 & 0.5000 & 0.4541 & 0.9251 & 289.24\\
    Figure~\ref{figure:7}(3) & \textit{SDv1-5}+\textit{Depth}+\textit{Davinci} & \textbf{0.9817} & 0.5582 & 0.7656 & 0.9281 & \textbf{0.9971} & 239.17\\
    Figure~\ref{figure:7}(4) & \textit{SDv1-5}+\textit{Depth}+\textit{Davinci}+\textit{MonaLisa} & 0.9087 & 0.5789 & 0.7500 & \textbf{0.9684} & 0.9963 & 233.21\\
    Figure~\ref{figure:7}(5) & \textit{SDv1-5}+\textit{Depth}+\textit{Davinci}+\textit{MonaLisa}+\textit{Leonardo} & 0.8279 & 0.5356 & 0.7656 & 0.7463 & 0.9689 & 220.40\\ 
    Figure~\ref{figure:7}(6) & \textit{SDv1-5}+\textit{Davinci} & 0.9184 & 0.3734 & 0.4843 & 0.5275 & 0.9200 & 265.01 \\
    Figure~\ref{figure:7}(7) & \textit{SDv1-5}+\textit{Davinci}+\textit{MonaLisa} & 0.9328 & 0.4591 & 0.6562  & 0.7235 & 0.9420 & 241.18\\
    Figure~\ref{figure:7}(8) & \textit{SDv1-5}+\textit{Davinci}+\textit{MonaLisa}+\textit{Leonardo} & 0.8876 & 0.4085 & 0.5937  & 0.6065 & 0.9458 & 275.92\\
    Figure~\ref{figure:7}(9) & \textit{SDv1-5}+\textit{MonaLisa} & 0.8778 & 0.4417 & 0.3593  & 0.1982 & 0.8251 & 336.24\\
    Figure~\ref{figure:7}(10) & \textit{SDv1-5}+\textit{Leonardo} & 0.8705 & 0.0964 & 0.6718  & 0.2673 & 0.9148 & 307.06\\
    Figure~\ref{figure:7}(11) & \textit{SDMv10} & 0.8759 & 0.0372 & 0.7031  & 0.2568 & 0.8837 & 320.48\\
    Figure~\ref{figure:7}(12) & \textit{SDMv10}+\textit{Depth}+\textit{Davinci} & 0.8898 & 0.5227 & 0.7343  & 0.6214 & 0.9819 & 209.06\\
    Figure~\ref{figure:7}(13) & \textit{SDMv10}+\textit{Depth}+\textit{Davinci}+\textit{MonaLisa} & 0.8876 & \textbf{0.5891} & 0.7656  & 0.5766 & 0.9834 & 212.13\\
    Figure~\ref{figure:7}(14) & \textit{SDMv10}+\textit{Depth}+\textit{Davinci}+\textit{MonaLisa}+\textit{Leonardo} & 0.8605 & 0.4096 & \textbf{0.8593} & 0.4481 & 0.9733 & \textbf{184.69}\\
    \bottomrule[1.5pt]
  \end{tabular}
  \begin{tablenotes}[flushleft]
		\item \textbf{Bold}: Best performance compared to all alliances.
            \item $\uparrow$:  Higher value corresponds to the generated image being more similar to the original image.
            \item $\downarrow$:  Lower value corresponds to the generated image being more similar to the original image.
     \end{tablenotes}
     
  \caption{Quantitative results under different Quantitative metrics. The Figure No. column corresponds to each generated Mona Lisa image in Figure~\ref{figure:7}. The Alliances column gives the model alliances used in each generated image(more quantitative data from the alliance, see Appendix~\ref{appendix}). The \textit{Cosine}$\sim$\textit{FID} column is the similarity score between each generated and original image under different quantification methods. }
  \label{tab:3}
\end{table*}
\begin{figure*}[t]
  \graphicspath{{./figures/}}
  \centering
  \includegraphics[width=1\linewidth, height=0.32\linewidth]{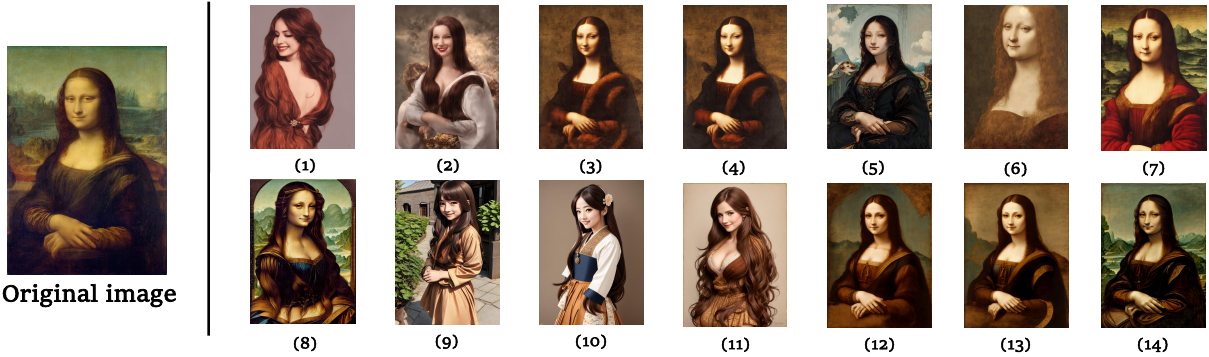}  
  \caption{The different models (e.g., \textit{SDMv10}, \textit{Depth}, \textit{Davinci}, etc.) are alliances to simulate the generation of Leonardo da Vinci's Mona Lisa images. In this example, we explore 30 alliances, including six models. We generate 100 batches of images for each alliance to explore the similarity between the images generated by different alliances and the original images.}
  \Description{}
  \label{figure:7}
\end{figure*}
\begin{figure*}[t]
  \centering
  \includegraphics[width=0.9\linewidth, height=0.315\linewidth]{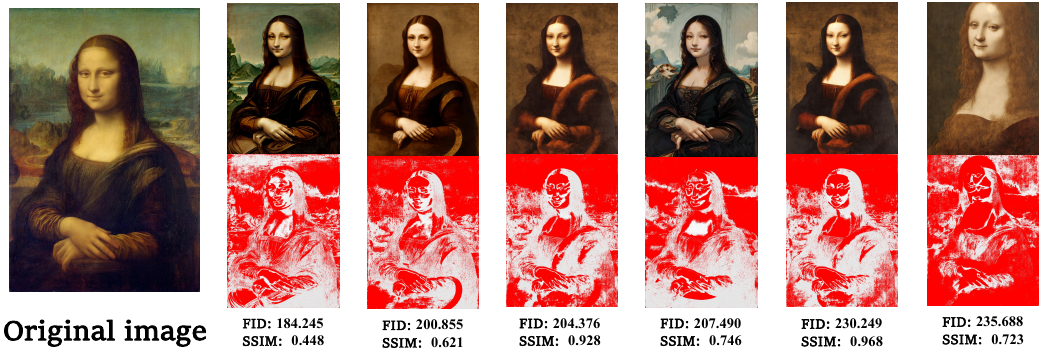}
  \caption{The pixel difference between the generated image and the original image, where darker color indicates larger differences, and lighter color indicates smaller differences.}
  \Description{}
  \label{figure:4}
\end{figure*}

\begin{itemize}
    \item \textbf{Property 1.} $ {\textstyle \sum_{z\in\mathcal{ M} }^{}} v(z)=U(\mathcal{ M})$. The Shapley values of all collaborators add up to the value of the grand coalition, ensuring that the total benefit is shared among them. 
    \item \textbf{Property 2.} For a player $z$,  if $U(\mathcal{L}) = U(\mathcal{L} \cup{z})$ holds for any  alliance $\mathcal{L}$, then $v(z)=0$.  This implies that if a model that participates in the diffusion workflow leaves the generated image unchanged from its absence, then this model does not affect the image generation and deserves zero contribution.
    \item \textbf{Property 3.} If player $z_i$ and player $z_j$ satisfy \(U(\mathcal{L} \cup {z_i}) = U(\mathcal{L} \cup z_j )\),  for any alliance $\mathcal{L}$ that does not include $z_i$ and $z_j$, then $v(z_i)=v(z_j)$. If multiple models contribute equally, they should be assigned the same contribution.
    \item \textbf{Property 4.} For two value functions $U$ and $U'$, we calculate the Shapley value of a player $z$ based on $U$, $U'$  and $U+U'$, denoted by $v(z)$, $v'(z)$ and $v''(z)$, respectively. Then, we have $v''(z)=v(z)+v'(z)$. If we combine two coalition games with value functions $U$ and $U'$, then the Shapley value of the combined coalition is equal to the sum of the Shapley values from each individual coalition. In our scenario, the image generation tasks based on different value functions are independent, and their contribution evaluations satisfy such additivity property.
\end{itemize}

Based on the above analysis, we demonstrate that our scenario theoretically fits the cooperative game model, and we can use the LOO and Shapley value methods to evaluate the model contribution in the diffusion workflow of image generation. We innovatively adopt the \textit{FID} as the value function and propose the \textit{FID}-Shapley algorithm to measure the contribution of each model alliance in AI image generation, as shown in Algorithm \ref{Algorithm:1}. The subsequent experiments show that \textit{FID}-Shapley can accurately reflect the contribution of each model to the AI-generated images, and also agree with the human  observation on the infringement. Models with high \textit{FID}-Shapley value have a more significant impact on the AI-generated images and are more likely to cause infringement issues, which provides guidance for AI-generated image users to pay extra attention to these models with high \textit{FID}-Shapley value in order to avoid copyright infringement in real-world.
\section{Experiments}
\label{experiment}
We conduct thorough experiments to ensure that the \texttt{CopyScope} framework can effectively solve the copyright traceability problem of AI-generated images.  
\subsection{Experimental Setup}
To investigate the specific infringement models involved in AI image generation, we use a diffusion workflow to generate Mona Lisa paintings, one of the classic artworks. We construct the AI image generation workflow using the models in Table \ref{tab:1}. We set \textit{SDv1-5} as the pipeline benchmark and assume its contribution to the generated image is zero. Then, we introduce \textit{SDMv10} as \texttt{Base model}, \textit{Depth} as  \texttt{ControlNet} model, \textit{Leonardo} as  \texttt{Lora} model, and "\textit{Davinci}" and "\textit{MonaLisa}" as two \texttt{Key Prompts}, and thus we obtain the models set $\mathcal{M} = \{$ \textit{SDMv10}, \textit{Depth}, \textit{Davinci}, "\textit{MonaLisa}", "\textit{Leonardo}" $ \} $. We keep the hyperparameters constant for all experiments, including image size, scale, sample step, etc. We construct a total number of $30$ diffusion workflows by different models alliances. With each diffusion workflow, we generate $100$ images and evaluate their contribution using their average \textit{FID} value. 

\subsection{Quantitative Metrics for Model Performance}
\label{quantifyexperiment}
In this experiment, we compare different quantitative metrics for model performance introduced in Section \ref{sec:similaritymetrics} and demonstrate that the \textit{FID} is the most effective metric for identifying the infringement models in AI image generation. Table~\ref{tab:3} presents the similarity values between the images generated by various diffusion workflows and the original images under different metrics. The \textit{Cosine} and \textit{RGB-SSIM} metrics give similar similarity values for different workflows, which implies that they are not sensitive to the model alliance. Thus, they are not informative for assessing the contribution of various models in the workflow. As shown in Figure~\ref{figure:7}, the \textit{Hist} and \textit{DHash} metrics, on the other hand, give similarity values that do not agree with the human perception of infringement. The \textit{Hist} metric only considers the color distribution of the images, and the \textit{DHash} metric only considers the hash fingerprint of the images. The \textit{SSIM} and \textit{FID} metrics have better discrimination ability on the similarity values between different generated images and original images, where \textit{SSIM} measures the structural similarity between images, reflected in higher similarity values for images with similar outlines, and \textit{FID} measures the similarity from more comprehensive image features, including outline, style, content, etc. As shown in Figure~\ref{figure:4}, under pixel-level evaluation, high \textit{FID} values correspond to overall similarity, while high \textit{SSIM} values correspond only to clear structural similarity. Therefore, \textit{FID} is more appropriate for tracing the infringement models in AI image generation.

\subsection{LOO vs. Shapley: Contribution Evaluation Experiment}

In this experiment, we compare the methods proposed in Section \ref{sec: evaluation_methods} for contribution evaluation, \textit{FID}-LOO and \textit{FID}-Shapley value.
\begin{figure}[t]
  \centering
  \includegraphics[width=1.0\linewidth, height=0.41\linewidth]{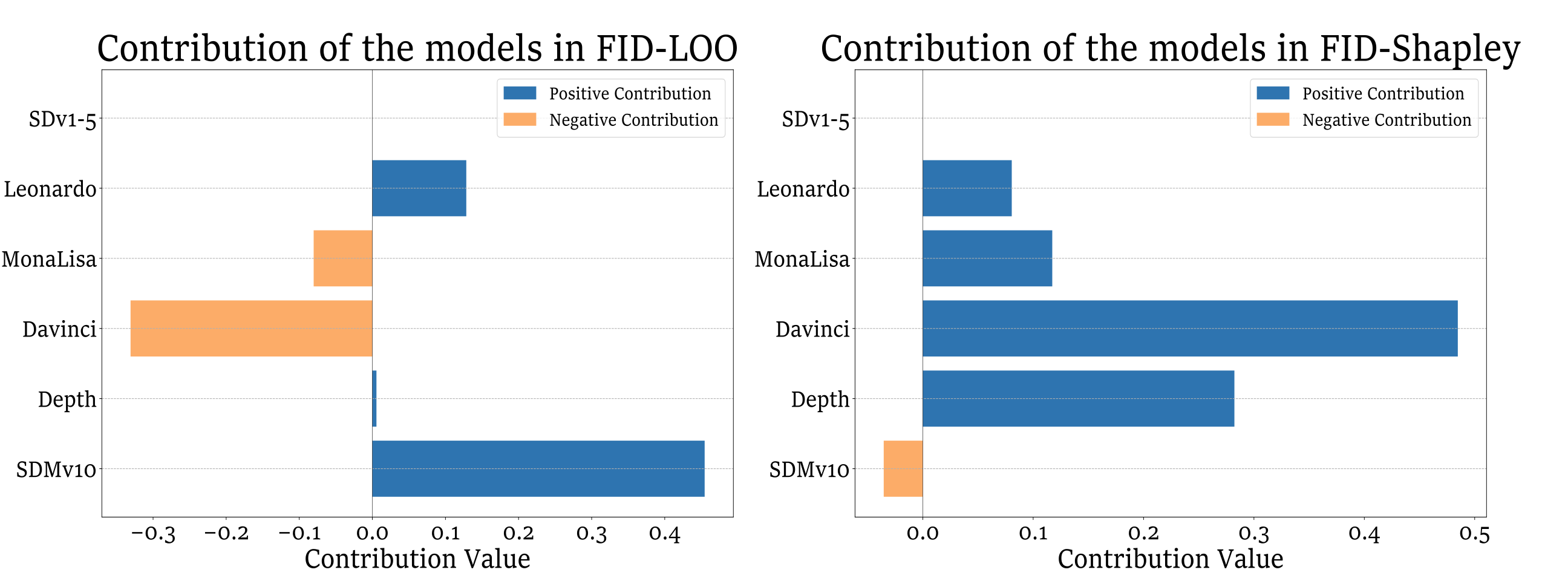}
  \caption{Comparison of contribution values under FID-LOO (left) and FID-Shapley (right) methods.
  }
  \label{figure:5}
\end{figure}
\begin{figure}[t]
  \centering
  \includegraphics[width=1.0\linewidth, height=1.37\linewidth]{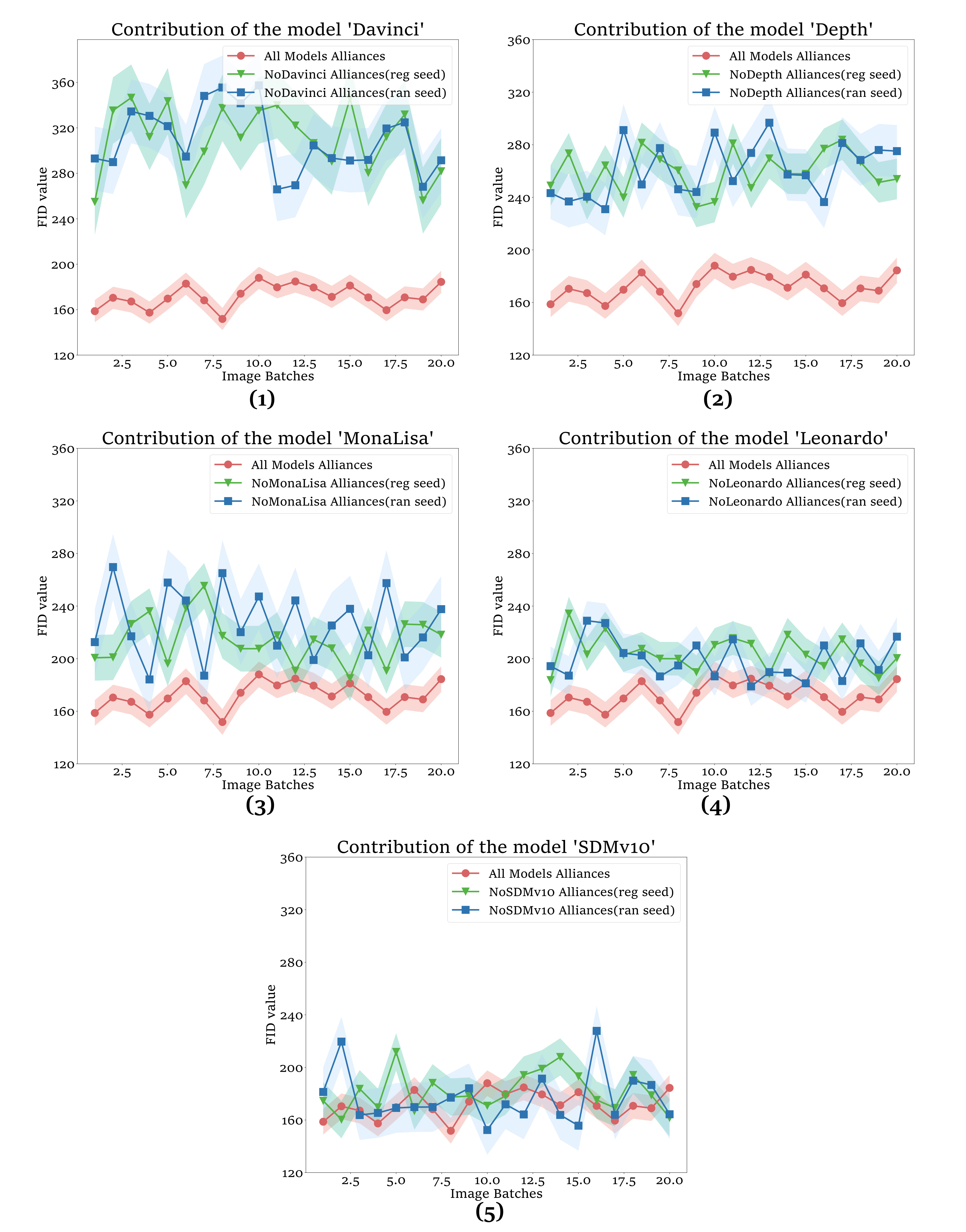}
  \caption{Ablation experiment. Corresponding models are eliminated in order, and the average \textit{FID} value of all alliance solutions composed of the remaining models is calculated. The greater the difference between the remaining alliances' \textit{FID} and the all-model alliance's \textit{FID}, the more influential the model is.}
  \label{figure:6}
\end{figure}

Figure \ref{figure:5} shows the normalized contribution of each model to the generation of AI image generation, calculated by the two methods. We observe that there are differences in the contribution evaluation between the two methods: under \textit{FID}-LOO, the contribution order is: $SDMv10>$"\textit{Leonardo}"$>Depth>$"\textit{MonaLisa}"$>Davinci$, while under FID-Shapley value, the contribution order is: $Davinci>Depth>$ "\textit{MonaLisa}"$>$"\textit{Leonardo}"$>SDMv10$. To verify which one of \textit{FID}-LOO and \textit{FID}-Shapley values provides a more accurate contribution evaluation in AI image generation, we further conduct ablation experiments. We drop out each model separately and calculate the average \textit{FID} value of the original image with the images generated by the diffusion models compromised by all subsets composed of the remaining models to measure the impact of the dropped model on the generated AI image generation. We repeat generating images of $100$ batches with regularized seed (reg seed: green line marked by triangles in Figure~ \ref{figure:6}) and random seed (ran seed: blue line marked by squares in Figure~ \ref{figure:6}). The red line marked by circles in Figure~ \ref{figure:6} shows the \textit{FID} of the AI images generated by the complete set of models and the original image. From Figure \ref{figure:6}, we observe that when dropping out \textit{Davinci}, it causes the largest deviation of \textit{FID} value, meaning that the similarity between the generated image and the original image significantly decreases after removing \textit{Davinci} model, which also means that \textit{Davinci} model has the largest contribution to the copyright of the generated AI image. Similarly, from Figure \ref{figure:6}, we can observe that the following important models are \textit{Depth}, \textit{"MonaLisa"} and \textit{"Leonardo"}, \textit{SDMv10}, respectively, which is consistent with the contribution evaluation results of \textit{FID}-Shapley value. The experiment proves that the \textit{FID}-Shapley value provides a more accurate and realistic contribution evaluation in AI image generation than \textit{FID}-LOO.

\section{Related Work}


\textbf{Data Attribution:} Some previous works traced copyright by evaluating the relationship between training input and generated output. For example, 
\citet{datta2016algorithmic} explored the impact of black-box machine learning model inputs on the algorithm by proposing a transparent mechanism.
\citet{pmlr-v202-park23c} introduced the method of using \textit{Tracing with the Randomly-projected After Kernel} (TRAK) to implement the data attribution problem of large-scale models.
\citet{wang2023evaluating} evaluated data attribution for text-to-image models through the “customization” method.
However, such methods were based on the premise of a known training dataset. As shown in Figure~\ref{figure:3}, in AI image generation, the user can generate images by using the selected model with some prompts. The specific images used to train the model were only known by the model provider. Therefore, tracing copyright from the perspective of training images is challenging to achieve in real-world practices.

\textbf{Fingerprint Traceability} Another sort of approach attempted to track the models by adding a fingerprint to the model. 
\citet{kim2023wouaf} modified the model according to each user’s unique digital fingerprint, so each text-to-image result was embedded with a unique fingerprint, and the model was traced through the fingerprint. \citet{8695364} experimented with several popular GAN architectures and datasets and demonstrated the existence of GAN fingerprints and their value for reliable forensic analyses.
\citet{10.5555/3618408.3619496} investigated the use of latent semantic dimensions as fingerprints, improved fingerprinting methods exhibit a significant tradeoff between robust attribution accuracy and generation quality, and enhanced the efficiency of fingerprint identification methods.
\citet{Fernandez_2023_ICCV} introduced an active strategy combining image watermarking and \textit{Latent Diffusion Models} (LDM) by fine-tuning the decoder of LDM and embedding watermarks in all the generated images. Although these work have studied various watermarking methods to achieve model traceability, these methods can only trace back to a chosen specific model in their experiments, without considering the interplay among models in the complex AI image generation task. 

\section{Conclusion}
To address the challenge of potential copyright infringement in AI-generated images, we have proposed a new framework called \texttt{CopyScope} that could identify different copyright infringement sources at the model level in the AI image generation process and evaluate their impact. We have proposed a \textit{FID}-based Shapley algorithm to assess the infringement contribution of each model in the diffusion workflow. Extensive results have demonstrated that our proposed \texttt{CopyScope} framework could effectively zoom in on the sources and quantify the impact of infringement models in AI image generation. Our work offers a promising solution for copyright traceability in AI image generation, which could also promote the legally compliant use of AI-generated content.

\normalem
\bibliographystyle{ACM-Reference-Format}
\bibliography{sample-authordraft}

\newpage
\begin{appendix}

\section{Appendix}
\label{appendix}

\begin{table}[htbp]
\centering
  \begin{tabular}{llccccccc}
    \toprule[1.5pt]
    \textbf{No.} & \textbf{Alliances }&  \textbf{\textit{Cosine}} $\uparrow$ & \textbf{\textit{Hist}}$\uparrow$ & \textbf{\textit{DHash}}$\uparrow$ & \textbf{\textit{SSIM}}$\uparrow$ & \textbf{\textit{RGB-SSIM}} $\uparrow$ & \textbf{\textit{FID}}$\downarrow$\\
    \midrule
     & \textit{SDv1-5} & 0.8802 & 0.1511 & 0.5468 & 0.2934 & 0.8424 & 310.18 \\
     & \textit{SDMv10} & 0.8759 & 0.0372 & 0.7031  & 0.2568 & 0.8837 & 320.48\\
    \hdashline
    1 & \textit{SDv1-5}+\textit{Depth} & 0.8927 & 0.3518 & 0.5000 & 0.4541 & 0.9251 & 289.24\\
    2 & \textit{SDv1-5}+\textit{Davinci} & 0.9184 & 0.3734 & 0.4843 & 0.5275 & 0.9200 & 265.01 \\
    3 & \textit{SDv1-5}+\textit{MonaLisa} & 0.8778 & 0.4417 & 0.3593  & 0.1982 & 0.8251 & 336.24\\
    4 & \textit{SDv1-5}+\textit{Leonardo} & 0.8705 & 0.0964 & 0.6718  & 0.2673 & 0.9148 & 307.06\\
    5 & \textit{SDMv10}+\textit{Depth} & 0.9063 & 0.4476 & 0.6250 & 0.5013 & 0.9876 & 292.57\\    
    6 & \textit{SDMv10}+\textit{Davinci} & 0.8512 & 0.2867 & 0.6250 & 0.2314 & 0.8694 & 216.07\\
    7 & \textit{SDMv10}+\textit{MonaLisa} & 0.8902 & 0.5384 & 0.6718 & 0.3225 & 0.9650 & 322.85 \\
    8 & \textit{SDMv10}+\textit{Leonardo} & 0.7085 & 0.5844 & 0.6562 & 0.3450 & 0.9557 & 240.05\\
    \hdashline
    9 & \textit{SDv1-5}+\textit{Depth}+\textit{Davinci} & \textbf{0.9817} & 0.5582 & 0.7656 & 0.9281 & \textbf{0.9971} & 239.17\\
    10 & \textit{SDv1-5}+\textit{Depth}+\textit{MonaLisa} & 0.8865 & 0.1114 & 0.6406 & 0.2734 & 0.8473 & 262.23\\
    11 & \textit{SDv1-5}+\textit{Depth}+\textit{Leonardo} & 0.7911 & 0.5074 & 0.5468 & 0.6238 & 0.9848 & 262.71\\
    12 & \textit{SDv1-5}+\textit{Davinci}+\textit{MonaLisa} & 0.9328 & 0.4591 & 0.6562  & 0.7235 & 0.9420 & 241.18\\
    13 & \textit{SDv1-5}+\textit{Davinci}+\textit{Leonardo} & 0.7543 & 0.4420 & 0.5156 & 0.8242 & 0.9864 & 260.68 \\
    14 & \textit{SDv1-5}+\textit{MonaLisa}+\textit{Leonardo} & 0.7730 & 0.5207 & 0.5000 & 0.2659 & 0.9279 & 332.45\\
    15 & \textit{SDMv10}+\textit{Depth}+\textit{Davinci} & 0.8898 & 0.5227 & 0.7343  & 0.6214 & 0.9819 & 209.06\\
    16 & \textit{SDMv10}+\textit{Depth}+\textit{MonaLisa} & 0.8798 & 0.5393 & 0.7343 & 0.4766 & 0.9910 & 246.01\\
    17 & \textit{SDMv10}+\textit{Depth}+\textit{Leonardo} & 0.8489 & 0.5735 & 0.7031 & 0.6685 & 0.9245 & 240.16\\  
    18 & \textit{SDMv10}+\textit{Davinci}+\textit{MonaLisa} & 0.8566 & 0.1206 & 0.6718 & 0.3229 & 0.8826 & 230.70 \\
    19 & \textit{SDMv10}+\textit{Davinci}+\textit{Leonardo} & 0.7897 & 0.5634 & 0.5468 & 0.2048 & 0.9280 & 301.20\\
    20 & \textit{SDMv10}+\textit{MonaLisa}+\textit{Leonardo} & 0.8248 & \textbf{0.6755} & 0.6406 & 0.1818 & 0.9254 & 310.06\\
    \hdashline
    21 & \textit{SDv1-5}+\textit{Depth}+\textit{Davinci}+\textit{MonaLisa} & 0.9087 & 0.5789 & 0.7500 & \textbf{0.9684} & 0.9963 & 233.21\\
    22 & \textit{SDv1-5}+\textit{Depth}+\textit{Davinci}+\textit{Leonardo} & 0.8156 & 0.5702 & 0.7187 & 0.8251 & 0.9528 & 227.45\\
    23 & \textit{SDv1-5}+\textit{Depth}+\textit{MonaLisa}+\textit{Leonardo} & 0.7977 & 0.6190 & 0.7187 & 0.4265 & 0.9564 & 217.41\\
    24 & \textit{SDv1-5}+\textit{Davinci}+\textit{MonaLisa}+\textit{Leonardo} & 0.8232 & 0.3436 & 0.5468 & 0.4178 & 0.8974 & 238.63\\
    25 & \textit{SDMv10}+\textit{Depth}+\textit{Davinci}+\textit{MonaLisa} & 0.8590 & 0.5709 & 0.7968 & 0.4528 & 0.9775 & 219.01 \\
    26 & \textit{SDMv10}+\textit{Depth}+\textit{Davinci}+\textit{Leonardo} & 0.8636 & 0.5636 & 0.8125 & 0.4807 & 0.9659 & 194.71\\
    27 & \textit{SDMv10}+\textit{Depth}+\textit{MonaLisa}+\textit{Leonardo} & 0.8142 & 0.5113 & 0.7500 & 0.4974 & 0.9315 & 240.71\\
    28 & \textit{SDMv10}+\textit{Davinci}+\textit{MonaLisa}+\textit{Leonardo} & 0.8081 & 0.6663 & 0.6250 & 0.2857 & 0.9558 & 248.54 \\
    \hdashline
    29 & \textit{SDv1-5}+\textit{Depth}+\textit{Davinci}+\textit{MonaLisa}+\textit{Leonardo} & 0.8279 & 0.5356 & 0.7656 & 0.7463 & 0.9689 & 220.40\\ 
    30 & \textit{SDMv10}+\textit{Depth}+\textit{Davinci}+\textit{MonaLisa}+\textit{Leonardo} & 0.8605 & 0.4096 & \textbf{0.8593} & 0.4481 & 0.9733 & \textbf{184.69}\\
    \bottomrule[1.5pt]
  \end{tabular}
     
  \caption{A full quantitative results under different metrics.}
  \label{tab:x}
\end{table}
\end{appendix}

\end{document}